\documentclass[letterpaper, 10pt, journal]{IEEEtran}
\IEEEoverridecommandlockouts

\usepackage[latin1]{inputenc}

\usepackage{paralist}
\usepackage{multirow}
\usepackage{cite}
\usepackage{amsmath,amssymb,amsfonts}
\usepackage{algorithm}
\usepackage{algorithmic}
\usepackage{graphicx}
\usepackage{textcomp}
\usepackage{float}

\usepackage{subfigure}
\usepackage{epstopdf}
\usepackage{cases}

\usepackage{verbatim}

\usepackage{amsmath}   

\usepackage{color,soul}
\soulregister\cite7 
\soulregister\citep7 
\soulregister\citet7 
\soulregister\ref7 
\soulregister\pageref7 
\setulcolor{bule} 
\setstcolor{yellow} 
\sethlcolor{yellow} 

\usepackage{bm}
\def\BibTeX{{\rm B\kern-.05em{\sc i\kern-.025em b}\kern-.08em
    T\kern-.1667em\lower.7ex\hbox{E}\kern-.125emX}}
\begin{document}

\title{
Directly Attention Loss Adjusted Prioritized Experience Replay}

\author{
	\IEEEauthorblockN{Zhuoying~Chen$^{1}$, Huiping~Li$^{1}$~\IEEEmembership{Senior Member,~IEEE}, Zhaoxu~Wang$^1$}
	
	\IEEEauthorblockA{$^1$ School of Marine Science and Technology,
		Northwestern Polytechnical University,  Xi'an, China, 710072}
	
	\IEEEauthorblockA{Email:czysmile@mail.nwpu.edu.cn, lihuiping@nwpu.edu.cn, wangzhaoxu@mail.nwpu.edu.cn}
}

\maketitle
\begin{abstract}
	
Prioritized Experience Replay (PER) enables the model to learn more about relatively important samples by artificially changing their accessed frequencies. However, this non-uniform sampling method shifts the state-action distribution that is originally used to estimate Q-value functions, which brings about the estimation deviation. In this article, an novel off policy reinforcement learning training framework called Directly Attention Loss Adjusted Prioritized Experience Replay (DALAP) is proposed, which can directly quantify the changed extent of the shifted distribution through Parallel Self-Attention network, so as to accurately compensate the error. In addition, a Priority-Encouragement mechanism is designed simultaneously to optimize the sample screening criterion, and further improve the training efficiency. In order to verify the effectiveness and generality of DALAP, we integrate it with the value-function based, the policy-gradient based and multi-agent reinforcement learning algorithm, respectively. The multiple groups of comparative experiments show that DALAP has the significant advantages of both improving the convergence rate and reducing the training variance.

\end{abstract}


\section{Introduction}

Experience Replay mechanism \cite{Lin1992}, as an general technical means of main stream off policy deep reinforcement learning algorithms, randomly selects a fixed number of experience transitions to provide training data for neural networks, which breaks the relevance between training samples. However, the equal probability sampling mode used in experience replay forces the agent to cost a lot of time to screen effective samples, which causes too much invalid exploration in the early stage of training. To resolve this problem, the Prioritized Experience Replay (PER) \cite{Schaul2016}  is proposed. PER takes the temporal difference (TD) error as a criterion for judging the prioritization of samples, and the frequency with a sample accessed is proportional to the absolute value of its TD error. The model training efficiency can be improved by learning the relatively important experience samples more frequently, and its performance has been verified in algorithms such as DQN \cite{Bu2020}, DDQN \cite{Tao2020}.

While PER improves the training speed, it also has the apparent shortcomings. On the one hand, the non-uniform sampling shifts the state-action distribution that is originally used to estimate the Q-value function, which brings about estimation deviation. On the other hand, the priority assignment form of PER is too singular, which is easy to cause overfitting problem\cite{Yue2023}.

From the estimation deviation aspect, the conventional PER introduces hyperparameter $\beta$ to adjust the importance sampling weight and compensate the esitimation error. $\beta$ increases linearly from the initial value $\beta_0$ to 1 with the increase of the training episodes \cite{Schaul2016}. However, this linear parameter tuning method will bring extra errors. Fujimoto \textit{et al}. \cite{Fujimoto2020} proposed a Loss Adjusted Prioritized Experience Replay (LAP) algorithm, which describes the loss function in segments. LAP suppresses the sensitivity of the quadratic Loss to outliers, and stops the further expansion of the error.
In order to solve the uncertainty caused by large TD error, Saglam \textit{et al}. \cite{Saglam2022} proposed an inverse sampling method to select samples with smaller TD errors for model training, which increases the training stability. Based on \cite{Schaul2016} \cite{Fujimoto2020}, we proposed an Attention Loss Adjusted Prioritized Experience Replay (ALAP) algorithm in the previous work \cite{Chen2023}, which utilizes Self-Attention network to construct a nonlinear mapping relationship between $\beta$ and training progress to further narrow the estimation deviation. Although the above works alleviate the error caused by PER to some extent, but they do not address the root cause of the error.

In terms of priority assignment, Gao \textit{et al}. \cite{Gao2021} constructed a new priority parameter as the criterion for screening  transitions, which is composed of reward and TD error. Gruslys \cite{Gruslys2017} takes advantage of the temporal locality of neighboring observations to set replay priority more efficiently. The work \cite{Sun2020} exploits attention network to calculate the similarity between the states in buffer and the current one, and the transition with more similar state is preferred to be selected. Brittain \textit{et al}. \cite{Brittain2020} proposed a algorithm called Prioritized Sequence Experience Replay (PSER), which increases the priority of the adjacent transitions before the goal (the transition with max TD error). The existing priority assignment methods are too singular in the setting of sample screening, which restricts the diversity of transtions. In this case, the overfitting and overestimation issues are easy to occur.

In order to solve the above problems, this paper proposes an novel algorithm called Directly Attention Loss Adjusted Prioritized Experience Replay (DALAP) that based on ALAP. Firstly, we theoretically prove the positive correlation between the esitimation error caused by shifted distribution and the hyperparameter $\beta$. After that, DALAP exploits the parallel Self-Attention network integrate with the Double-Sampling mechamism \cite{Chen2023} to calculate the distribution similarity of random uniform sampling (RUS) and priority based samping (PS), simultaneously. On this basis, the Similarity-Increment (SI) generated by PS is solved by making the difference of the sample similarity between the two sampling methods. The alteration degree of the shifted distribution is accurately quantified by SI, so as to fit a more accurate $\beta$ to eliminate the error from the root. 

To screen out samples that are more worthy of learning, we propose an Priority-Encouragement (PE) mechanism, which affirm the neighboring transitions before the goal are also possess some learning value. We increase the frequency they are visited by raising their priority, and the priority growth of each transition decays as it moves away from goal under the effect of the decay coefficient $\rho$. In addition, we also design a greedy priority sequence, which makes $\rho$ itself decay to 0 with the advance of training. In other words, the role of Priority-Encouragement will decrease with the progress of training. At the same time, in order to reduce the computational load, we weed out the non-essential items that have no learning value from PE.

As a general training framework for reinforcement learning, DALAP resolves the shortcomings of the existing PER variants in both the estimation deviation and the priority assignment aspects, simultaneously. In order to verify the superiority of DALAP, we integrate it with DQN \cite{Mnih2015} \cite{Sharma2021}, DDPG \cite{Timothy2016} \cite{Yang2022} and MADDPG \cite{Lowe2017}, respectively, and beat ALAP algorithm in three different environments. The main contribution of this study are shown as follow:

\begin{itemize}
\item The theoretical proof of the positive correlation between the shifted distribution caused estimation error and the hyperparameter $\beta$ is proposed.

\item A parallel Selt-Attention network is proposed to simultaneously compute the similarity of the sample distributions for two parallel sampling methods. On this basis, the Similarity-Increment due to PS is calculated, which can directly quantify the change degree of the shifted distribution, and fit a more accurate $\beta$.

\item A Priority-Encouragement mechanism is proposed, which utilizes a greedy priority sequence to improve the priority of the adjacent transition before the goal, and removes the unnecessary items with no learning value from the sequence to reduce the computational load.

\end{itemize}

In the rest of the article: Section \ref{PRELIMINARIES} introduces the basic knowledge and makes a statement about the problem. Section \ref{section:DALAP} describes the design process of the DALAP algorithm in detail. Section \ref{section:experiment} provides the comparative experiment results and analysis. Finally, Section \ref{conclusion} draws a conclsion of the whole work.

\section{PRELIMINARIES}\label{PRELIMINARIES}

\subsection{Prioritized Experience Replay}\label{PER}
As a non-uniform sampling strategy, PER artificially changes the sampled frequency of the experience transitions, so that the model can learn the important one more frequently. In PER, the sampled probability of each sample $i$ is proportional to its TD error:
\begin{equation}\label{prioritiy_distribution_function}
	\begin{aligned}
		P(i)=\frac{p_i^{\alpha}}{\sum_{j}p_j^{\alpha}},
	\end{aligned}
\end{equation}
where $p_i=\left|\delta(i)\right|+\epsilon$ is the priority of transition $i$. The hyperparameter $\alpha $ is applied to smooth out the extremes, and a minimal positive constant $\epsilon$ is added to make sure that the sampled probability of each transition is greater than 0.

Considering that PER shifts the state-action distribution, the importance sampling weight $w(i)$ is introduced to correct it:

\begin{equation}\label{importance sample weights}
	\begin{aligned}
	w(i) = \left( \frac{1}{N}\cdot\frac{1}{P(i)} \right)^\beta,
	\end{aligned}
\end{equation}
and PER normalizes weights by $1/max_jw_j$ for stability reason:
\begin{equation}\label{normalize weights}
	\begin{aligned}
		\bar{w}(i)=\frac{w(i)}{\max_{j}w(j)},
	\end{aligned}
\end{equation}

\begin{equation}\label{PERloss}
	\begin{aligned}
		L_{PER}(\delta(i))=\bar{w}(i)L(\delta(i)).
	\end{aligned}
\end{equation}

Note that $\bar{w}(i)$ is embedded in formula (\ref{PERloss}) when updating the network. The hyperparameter $\beta$ increases linearly from $\beta_0$ to 1 as the episode increase, and the deviation caused by shifted distribution will completely compensated when $\beta$ reaches 1.

\subsection{Loss Adjust Prioritized Experienced Replay}\label{LAP}

PER uses the Mean Square Error (MSE) as the Loss to update the network, and its sensitivity to outliers is easy to further amplify the deviation. LAP exploits the segemental loss function to deal with TD errors of different sizes:

\begin{equation}\label{huber}
	L_{Huber}(\delta(i))=
	\begin{cases}
		0.5\delta(i)^2 & 	\left|\delta(i)\right|\leq 1,\\
		\left|\delta(i)\right| & otherwise,\\
	\end{cases}
\end{equation}
and combined with priority clipping scheme to alleviate the deviation:
\begin{equation}\label{clipping}
	\begin{aligned}
		P(i)=\frac{max(\left|\delta(i)\right|^\alpha,1)}{\sum_{j}max(\left|\delta(j)\right|^\alpha,1)}.
	\end{aligned}
\end{equation}

It can be seen from fomular (\ref{huber}) that LAP utilizes different loss functions to update the network in the face of different sizes of TD errors. The MSE is applied as the loss when the absolute value of TD error is less than or equal to 1, and MAE is used in other cases to suppress the sensitivity to outliers. It is worth noting that when MSE is executed, the priority of transition $i$ is clipped to 1, which means the uniform sampling is carried out with a sampled probability of $\frac{1}{N}$ according to formular (\ref{clipping}).

\subsection{Attention Loss Adjusted Prioritized Experience Replay}\label{ALAP}

PER ignores the bias generated during the training process, and only focuses on the unbiasedness of the convergence stage. It believes that the error compensation strength should increase with the convergence process (training progress), so PER makes $\beta$ increases linearly with the episodes increase. However, the training progress is not uniformly distributed on the scale of the episodes, so a linearly varying $\beta$ may introduce additional errors. ALAP utilizes the similarity of transitions to quantify the training progress, and proposes an improved Self-Attention network (SAN), which can calculate the simularity of the distribution in buffer:
\begin{equation}\label{self-attention}
	\begin{aligned}
		Self-Attention(Q,K)=softmax(\frac{Q \bullet K^T}{\sqrt{d_k}}),
	\end{aligned}
\end{equation}
\begin{equation}\label{projection}
	\begin{aligned}
		Q \bullet K^T= \sum_{i=1}^{m}\frac{(q_i\cdot k_i)}{\left|k_i\right|}
		.
	\end{aligned}
\end{equation}

The input of the Self-Attention network is the mini-batch sized state-action pair sequence $X = [(s_1,a_1),..., (s_m,a_m)]$, and the output is the similarity of the internal elements of $X$. In the formula (\ref{self-attention}) and (\ref{projection}), $Q=XW_Q$ , $W_Q$ represents the initial weight matrix of $Q$, and the elements in $Q$ are randomly rearranged (by shuffle operation) to get $K=shuffle(Q)$. According to formular (\ref{projection}), ALAP computes the similarity of the corresponding vectors $q_i$ and $k_i$ through vector projection, and estimates the similarity between $Q$ and $K$ by summing all the projected values, which represented by $Q \bullet K^T$.

In addition, since PS uses fixed criterion to screen out samples, the sample distribution in $X$ does not reflect the original one in the buffer $D$. Therefore, ALAP proposes a Double-Sampling (DS) mechanism based on mirror buffer $D^*$, where the algorithm simultaneously adopts two parallel sampling methods: RUS is responsible for providing inputs for Self-attention, and PS provides high-quality training samples for the model. The mirror buffer has the same data distribution as the original one, which ensures the stability of the parallel sampling. 

\subsection{Prioritized Sequence Experience Replay}\label{PSER}

PSER records the transition with max TD error in buffer as the goal $G$, and gives some priority growth to each neighboring transions before $G$:
\begin{equation}\label{pser}
\begin{aligned}
 & p_{n-1}=min(p_n \rho + p_{n-1}, \max\limits_n p_n) \\ 
 & p_{n-2}=min(p_n \rho^2 + p_{n-2},\max\limits_n p_n) \\
 & \qquad \qquad \qquad \ldots \\
 & p_{n-i}=min(p_n\rho^i + p_{n-i},\max\limits_n p_n),
\end{aligned}
\end{equation}
Where $\rho$ represents the decay coefficient, $p_n$ is the priority of $G$, and $P_{n-i}$ is the priority of the transition that is located $i$ steps before $G$ in the sample sequence.
Note that the priority growth $p_n\rho^i$ of each transition will decay as their distance from $G$ increases. To avoid the computation waste, the gorwth dacay has a threshold of $1\%$ $p_n$, after which the decay will stop. PSER set a window W to realize this function by limiting the times of decay:

\begin{equation}\label{window}
	\begin{aligned}
		p_n \rho^W \leqslant 0.01p_n, W\leqslant \frac{ln0.01}{ln\rho}
	\end{aligned}
\end{equation}

\section{Directly Attention Loss Adjusted Prioritized Experience Replay}\label{section:DALAP}

This section will describe the design process of DALAP algorithm from three parts: theoretical foundation, Parallel Self-Attention network, and Priority-Encouragement mechanism.

\subsection{Theoretical Foundation}\label{proof}

We start our algorithm design by establishing the theoretical foundation of the DALAP training framework, which proves the positive correlation between the estimation error and the hyperparameter $\beta$. We first introduce the \textbf{Threom 1}, which identified by Saglam \textit{et al}. (2022) to support our argument.

\noindent\textbf{Threom 1.} If $\delta_{\theta}$ is the TD error that relevant with the $Q_{\theta}$ network, then there exists a transition tuple $\tau _t= (s_t,a_t,r_t,s_{t+1})$ with $\delta_{\theta} \neq 0$ so that if the magnitude of TD error on $\tau _t$ increase, the magnitude of estimation error of $Q_{\theta}$ on at least  $\tau _t$ or $\tau _{t+1}$ will increase as well:
\begin{equation}\label{theorem1}
	\begin{aligned}
		\left|\delta(\tau _t)\right| \propto \underbrace{\left|Q_{\theta}(s_i,a_i)-Q^*(s_i,a_i)\right|}_{\varepsilon}; i=t \vee t+1,
	\end{aligned}
\end{equation}
where $\varepsilon$ stands for the esitimation error. $Q_{\theta}(s_i,a_i)$ and $Q^*(s_i,a_i)$ are estimated and optimal value under policy $\pi$, respectively. We now follow the \textbf{Threom 1} to prove our conclusion \textbf{Lamma 1}. For ease of description, we use $\Uparrow$ and $\Downarrow$ to define the increase or decrease of values in the whole process of the proof, respectively. 

\noindent\textbf{Lamma 1.} The esitimation error caused by shifted distribution $\varepsilon_t$ is positively correlated with the hyperparameter $\beta$, which is applied to regulate the error correction strength:

\begin{equation}\label{lamma1}
	\begin{aligned}
	\varepsilon_t \propto	\left|\delta(\tau _t)\right|	 \propto \beta.
	\end{aligned}
\end{equation}
\noindent \textit{proof}. We should first clarify that in PER, $\varepsilon_t$ is the main part of $\varepsilon$, so we ignore the rest of $\varepsilon$ that is much smaller than $\varepsilon_t$, and it's easy to obtain $\varepsilon_t 	= \varepsilon  \propto \left|\delta(\tau _t)\right| $ according to formaular (\ref{theorem1}). Then, bringing formular (\ref{importance sample weights}) into (\ref{normalize weights}) yields: 
\begin{equation}\label{normalize expansion}
	\begin{aligned}
		\bar{w}(i)= \frac{\left( \frac{1}{N}\cdot\frac{1}{P(i)} \right)^\beta}{N\cdot \left( \frac{1}{N}\cdot\frac{1}{min(P(j))} \right)^\beta  } =\frac{1}{N} \cdot \underbrace{\left(\frac{min(P(j))}{P(i)}\right)^\beta}_{\textit{f}(\beta)},
	\end{aligned}
\end{equation}
where $P(i)$ is a non-zero probability distribution according to formular (\ref{prioritiy_distribution_function}), and it is obvious that $\frac{min(P(j))}{P(i)}$ $\in$ (0,1).

When 	$\varepsilon_t$ $\Uparrow$, the $\left|\delta(\tau _t)\right|$ will $\Uparrow$ as well, which lead the loss $L(\delta(\tau _t))$ $\Uparrow$ enventually. In addition, we've already known that $L_{PER}(\delta(\tau _t))=\bar{w}(t)L(\delta(\tau _t))$ from formular (\ref{PERloss}), and our goal is to minimize the loss $L_{PER}(\delta(\tau _t))$ as the optimization direction. Since $\left|\delta(\tau _t)\right|$ is $\Uparrow$ at current step, the only way to 
$\Downarrow$ $L_{PER}(\delta(\tau _t))$ is $\Downarrow$ the $\bar{w}(i)$. We can easily deduce that $\bar{w}(i)$ $\propto$ $\textit{f}(\beta)$ as $\frac{1}{N}$ is a positive constant. Considering that $\frac{min(P(j))}{P(i)}$ $\in$ (0,1), the  $\textit{f}(\beta)$ is the exponential function with $\frac{min(P(j))}{P(i)}$ as the base, which decreases monotonously within the range of $\beta$. That is to say, you need to $\Uparrow$ $\beta$ to meet the requirement of $\Downarrow$ both $\bar{w}(i)$ and $\textit{f}(\beta)$. Therefore, $\beta$ $\Uparrow$ as the $\left|\delta(\tau _t)\right|$ $\Uparrow$. So far, we demonstrate the positive correlation between $\left|\delta(\tau _t)\right|$ and $\beta$: $\left|\delta(\tau _t)\right|$ $\propto$ $\beta$.

To sum up, we draw our conclusion $\varepsilon_t \propto	\left|\delta(\tau _t)\right|	 \propto \beta$ in the basis of $\varepsilon_t = \varepsilon  \propto \left|\delta(\tau _t)\right| $ and $\left|\delta(\tau _t)\right|$ $\propto$ $\beta$. The key point of DALAP is to find a more accurate $\beta$ to compensate the esitimation error, and this could be figure out if we can quantify the $\varepsilon_t$.

\subsection{Design of Parallel Self-Attention Network}\label{PSA}
ALAP follows PER's idea of error compensation, which argues that the  compensation strength should increase with the training progress. However, this approach only emphasizes the unbiasedness at the convergence stage and ignores the error effect in the remaining part. What's more, ALAP is unable to directly calculate the extent to which PER shifts the state-action distribution, and does not address the root cause of the problem.

To resolve this problem, a Parallel Self-Attention Network (PSAN) is proposed on the basis of the Self-Attention network (SAN) that is applied in ALAP \cite{Chen2023}. The PSAN can directly quantify the influence of PER on the distribution to calculate $\varepsilon_t$, and fit a more accurate $\beta$ according to \textbf{Lamma 1.}.

As training progresses, the agent tend to exploit rather than explore. In other words, given a state, the agent may choose roughly the same action, at which point the transitions generated by the agent interacting with the environment will gradually become similar. We denote this training-generated similarity as $I_t$. When the input of SAN is the transitons provided by RUS, this random uniform sampling method ensures the consistency of sample distribution between mini-batch and experience pool, so the output is $I_t$ in this case. When PS is used as the input to SAN, the output of SAN at this point is similarity $I_p$. It is worth noting that $I_p$ consists of two parts, one is the sample similarity brought by training itself ($I_t$); the other is the Similarity-Increment brought by PS screening samples with a fixed criterion, which we denote as $\Delta_i$. It is obvious that when we remove the similarity $I_t$ that generated by training in $I_p$, the remain part is the Similarity-Increment $\Delta_i$ due to the use of PS: 
\begin{equation}\label{SI}
	\begin{aligned}
	\Delta_i = I_p-I_t,
	\end{aligned}
\end{equation}
where the Similarity-Increment $\Delta_i$ quantify the extent to which PER changes the state-action distribution, and this change is the root cause of the estimation error $\varepsilon_t$. As a result, we obtain:
\begin{equation}\label{SI_BETA}
	\begin{aligned}
		\Delta_i \propto \varepsilon_t \propto \beta,
	\end{aligned}
\end{equation}
and we can get the exact $\beta$ after normalizing $\Delta_i$ according to formular (\ref{SI_BETA}).

\begin{figure}[htbp] 
	\begin{minipage}{1\linewidth}
		\centering
		\includegraphics[width=0.8\textwidth]{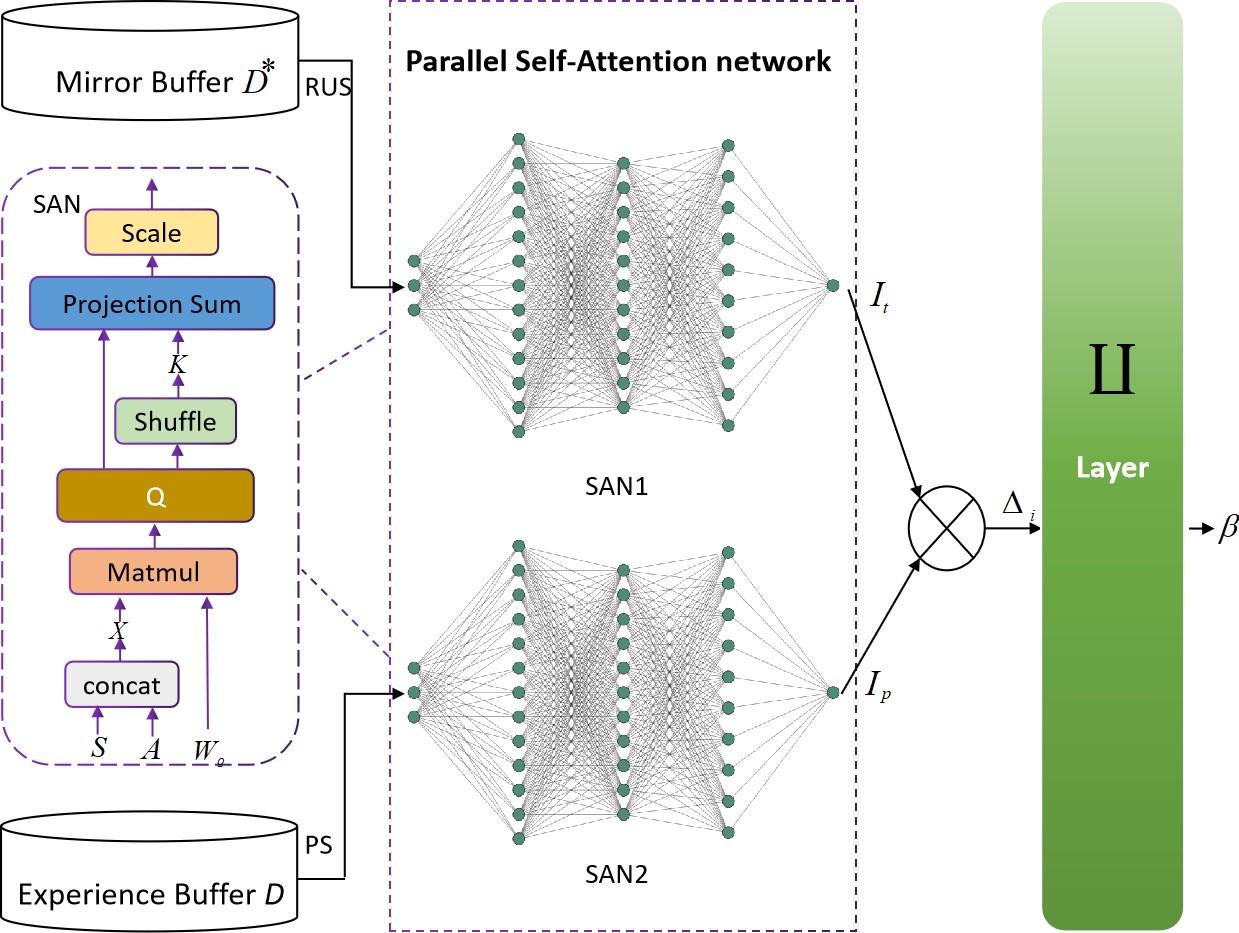}
	\end{minipage}
	\caption{Parallel Self-Attention Network.}
	\label{FigPSAN}
\end{figure}

Clearly, if we know $I_p$ and $I_t$, we can calculate $\Delta_i$ according to formula (\ref{SI}). However, the SAN adopted by ALAP can only compute one of them individually, which is the reason we design PSAN. PSAN expands the network structure of SAN by adapting the single-input single-output system (SISO) to multiple-input multiple-output (MIMO). PSAN has two parallel SAN networks, which can simultaneously receive two different data sources (RUS, PS) provided by the Double-Sampling mechanism, and compute the $I_p$ and $I_t$, respectively. The schematic diagram of PSAN is shown in Fig.\ref{FigPSAN}.  

From Fig.\ref{FigPSAN} we can see, the PSAN contains two SAN that share the same network structure, the hyperparameter $\beta$ can be obtained after entering $\Delta_i$ into the normalization layer $\coprod$.
\subsection{Design of Priority-Encouragement Mechanism}\label{LAP}

The preceeding paragraphs \ref{proof} and \ref{PSA} improve PER mainly in terms of reducing estimation deviation, while this subsection will improve PER with respect to priority assignment.

PS adopts a singular sample selection criterion, and is approximately greedy of selecting the sample with the largest TD error for training, which makes training samples too single and easy to bring overfitting. When facing the sparse reward \cite{Zhang2023} \cite{Zhao2023} environment, since the absolute value of TD error corresponding to most transitions is very small at the beginning of training, it will aggravate the greedy characteristic of PS on sample selection if the mini-batch size is small as well.

In order to address this problem, we propose a Priority-Encouragement (PE) mechanism, which enriches sample diversity by adding priority to samples with small TD error but still have some value to learn. PE follows the basic conclusions and variable symbol settings of PSER in \ref{PSER}. It believe that the adjacent transitions before reaching the goal $G$ also have some learning value, even if their TD errors are small. Therefore, we give these transitions some priority growth, which will decay as their distances from $G$ increase under the effect of the decay coefficient $\rho$:

\begin{equation}\label{pser2}
	\begin{aligned}
		& p_{n-1}=min(p_n \rho + p_{n-1},p_n) \\ 
		& p_{n-2}=min(p_n \rho^2 + p_{n-2},p_n) \\
		& \qquad \qquad \qquad \ldots \\
		& p_{n-i}=min(p_n\rho^i + p_{n-i},p_n),
	\end{aligned}
\end{equation}
where $p_n$ denotes the priority of $G$. Different from PSER, PE brodens the range of $G$. PSER only labels the transition with the highest priority in buffer as $G$, which limits the diversity of the training samples. In addition, in the sparse reward environment, the initial TD error of most transitions in buffer is close to 0, which is because the reward value of most states are 0, and the estimated $Q$ function based on reward expectation is also close to 0. Therefore, it can be seen that the TD error consisting of $Q$ and the instant reward $r_t$ is close to 0 as well, according to formular (\ref{Delta_expansion}):

\begin{equation}\label{Delta_expansion}
	\begin{aligned}
		\delta(\tau _t) = r_t + \gamma Q_{\theta}(s_t+1,a_t+1)-Q_{\theta}(s_t,a_t).
	\end{aligned}
\end{equation}

In this condition, the transitions in the buffer have almost the same priority, at which time the priority of $G$ ($\max\limits_n p_n$), according to formular (\ref{pser}) has no significant advantage over other transitions. In other words, $G$, at this moment, is probably not the most worthwhile sample to learn, and the priority propagation based on $G$ alone is likely to cause an overestimation problem. PE builds on PSER by extending the scope of $G$ to all transitions in the mini-batch excluding the item where the minimum priority is located. That is, we believe that almost all the samples screened out by PS can be used as the goal $G$. One point to emphasize here is that PER constructs a non-zero probability distribution rely on $\epsilon$, which gives all samples a chance to be selected, according to formular (\ref{prioritiy_distribution_function}). For samples with small TD errors, this chance is small, but does exists. Obviously, such samples are not what we need for $G$, and the transitions adjacent to them should not be assigned additional priority. As a result, we avoid this chance by excluding the smallest priority term in the mini-batch from $G$.

To relieve the computation load, we follow the window $W$ exploited by PSER to limit the number of decays. In addition, considering that the agent's choice between exploration and exploitation will gradually favour the latter as the training progresses, and its need for sample diversity will diminish as the number of exploration decreases. Hence, we propose a greedy encourage strategy that progressively weakens the strength of priority encouragement through a decreasing decay coefficient $\rho$:

\begin{equation}\label{rho}
	\begin{aligned}
		\rho=\rho_0(1-e/e_{total}),
	\end{aligned}
\end{equation}
where $\rho_o$ represents the initial value of $\rho$. $e$ and $e_{total}$ denote current and total episodes, respectively. At this point, we have completed the design of DALAP, and its running flow can be found in \textbf{Algorithm 1}.

\begin{algorithm}[htbp] 
	\caption{DALAP Algorithm}
	\begin{algorithmic}[1]
		\renewcommand{\algorithmicrequire}{\textbf{Input:}}
		\renewcommand{\algorithmicensure}{\textbf{Output:}}
		\REQUIRE mini-batch  $m$, step-size $\sigma$, replay period $K$, buffer sotrage $N$, exponent $\alpha$ and $\beta$, budget $T$, tiny positive contant $\epsilon$.
		\STATE Initialize replay buffer $D = \emptyset$, mirror replay buffer $D^* = \emptyset$, $p_1=1$, $\Delta=0$
		\STATE Observe environment state $s_0$ and choose action $a_0$
		\FOR{t=1 \TO $T$}
		\STATE Observe $s_t$, $r_t$, $\gamma_t$
		\STATE Store transition $(s_{t-1},a_{t-1},r_t,\gamma_t,s_t)$ in $D$
		\STATE Store transition $(s_{t-1},a_{t-1},r_t,\gamma_t,s_t)$ in $D^*$
		\IF {$t > K$}
		\FOR {t=1 \TO $K$}
		\STATE Sample transition $j \sim P(j)=p_j^\alpha/\sum_{i}p_i^\alpha$

		\STATE Obtain transitions $S$ with mini-batch $m$ from $D$ by PS for model training
		\STATE Obtain goal transtions $G$ after removing the minimum priority term from $S$
		\FOR {every $G_i$ in $G$}
		\STATE Assign priority growth to the transitions before $G_i$ according to PE
		\ENDFOR
		\STATE Obtain transitions $S^{*}$ with mini-batch $m$ from $D^*$ by RUS
		\STATE Obtain $\beta$ after inputing $S$ and $S^{*}$ into PSAN 
		\STATE Compute TD-error $\delta_j$ and update transition priority $p_j\gets \left|\delta(j)\right| + \epsilon$ 
		\STATE Compute importance sampling weight $\bar{w}(i)=(N\cdot P(j))^{-\beta}/\max_{i}w(i)$
		\STATE Compute weight-change $\Delta \gets \Delta +\bar{w}(j)\cdot \delta(j) \cdot \nabla_{\theta}Q(s_{t-1},a_{t-1})$
		\ENDFOR
		\STATE Update weights $\theta \gets \theta+\sigma \cdot \Delta$, reset $\Delta=0$
		\STATE Copy parametes to target network $\theta_{target} \gets \theta$
		\ENDIF
		\STATE Choose action $a(t) \sim \pi_{\theta}(s_t)$
		\ENDFOR

	\end{algorithmic}
\end{algorithm}

\section{Experiment}\label{section:experiment}

In order to demonstrate the outstanding performance of DALAP, we embed it into DQN, DDPG, and MADDPG, respectively, and compare it with ALAP in the corresponding \textbf{OPENAIgym} \cite{Brockman2016} \cite{Rezazadeh2020} environments such as \textbf{cartpole-v0}, \textbf{simple}, \textbf{simple$\_$tag}, etc. In addition, to further demonstrate the stability of DALAP, and to exclude the interference of some accidental results. The comparison experiments will be carried out under different mini-batch configurations in each environment. In each case, we have run 20 tests, and set a 50$\%$ confidence interval to draw the confidence band, which can describe the magnitude of training variance for each algorithm.

\subsection{Results with DQN}\label{DQNresult}

In \textbf{cartpole-v0}, DQN will be used as a carrier for each training framework, and we compare DALAP, ALAP, LAP, and PER after integrating them with DQN, respectively. All these training frameworks only change the training mode of the model, and do not modify its basic network structure. At the same time, to ensure that the training method is the only difference in the comparison test, all the algorithms share the uniform hyperparameter configuration. During training, these algorithms will excute different sampling manners to draw mini-batch samples from the corresponding experience pool of volume $2\times10^4$. The model is consturcted by 3-layer ReLU MLP with 24 units per layer. The total episodes of training, the maximum simulation step, and the maximum reward value are all 200. Furthermore, the adam optimizer will be exploited for network updating with learning rate and discount factor of 0.001 and 0.99, respectively.

\begin{figure*}[!t]
	\centering
	\subfigure[batchsize=32]{
		\includegraphics[scale=0.355]{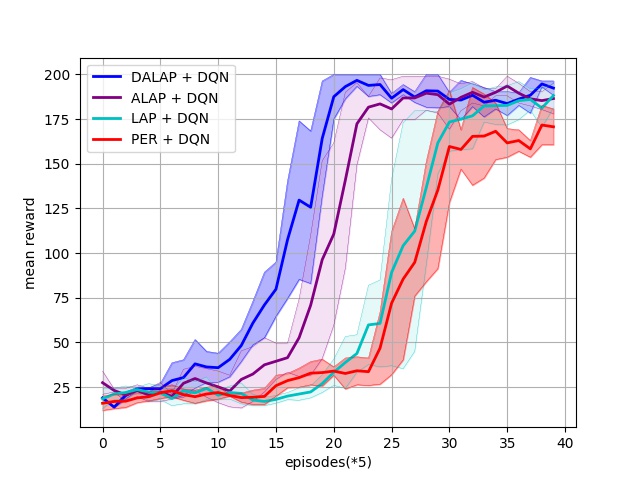}}
	\subfigure[batchsize=64]{
		\includegraphics[scale=0.355]{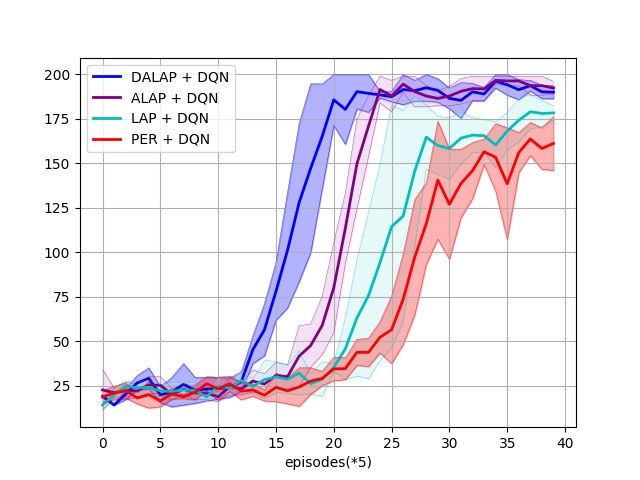}}
	\subfigure[batchsize=128]{
		\includegraphics[scale=0.355]{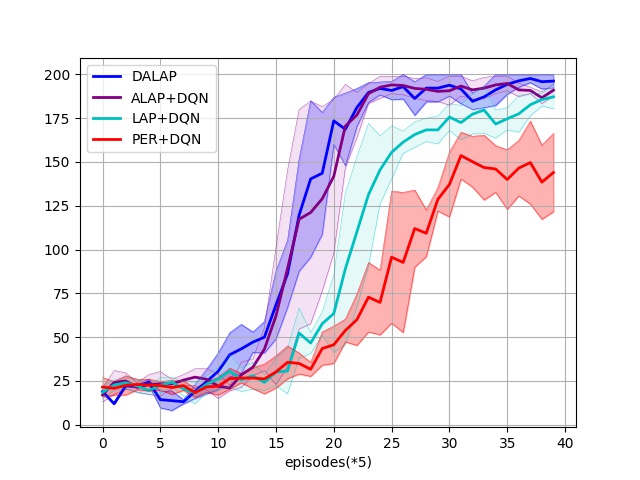}}
	\caption{Mean rewards of different training frameworks integrated with DQN.}
	\label{DQN}
\end{figure*}

Fig.\ref{DQN} describes the average reward curves of different training frameworks combined with DQN, and it is very clear to see that DALAP converges faster and has higher steady-state rewards, which gives it a better performance compared to other algorithms. DALAP has the smallest confidence bands throughout the training process under different mini-batch configurations, which suggests its excellent stability. When mini-batch=128, DALAP has the equivalent convergence speed and steady-state rewards as ALAP, but its confidence band is much smaller than ALAP, indicating that DALAP possesses a smaller training variance at this point.

\subsection{Results with DDPG}\label{DDPGresult}

Similar to \textbf{cartpole-v0}, DDPG only plays the role of a training framework carrier in \textbf{simple}. Considering the increased complexity of the environment, we adjusted the sample volume of the experience pools $D$ and $D^*$ to $10^6$, and they will provide training samples for the 3-layer ReLU MLP network that accommodates 64 neurons per layer. During training, the agent rewarded for approaching landmark. The total episodes of training are 2000, the maximum simulation step is 25. The learning rate and the discount factor are 0.001 and 0.95, respectively.

\begin{figure*}[!t]
	\centering
	\subfigure[batchsize=32]{
		\includegraphics[scale=0.355]{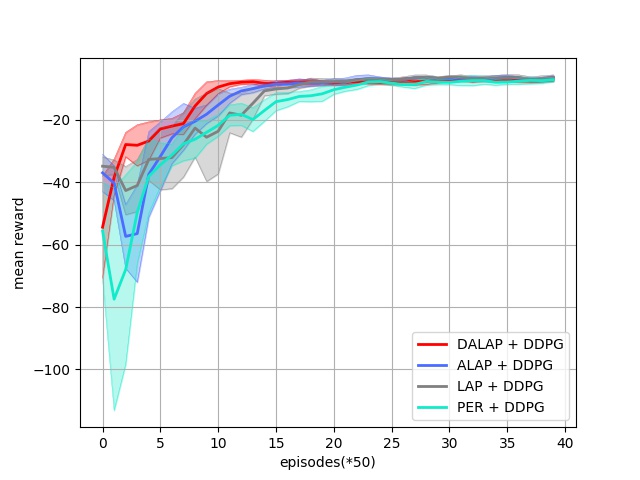}}
	\subfigure[batchsize=64]{
		\includegraphics[scale=0.355]{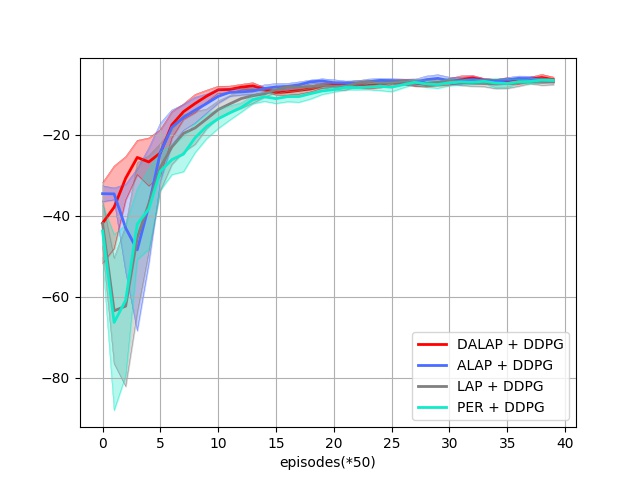}}
	\subfigure[batchsize=128]{
		\includegraphics[scale=0.355]{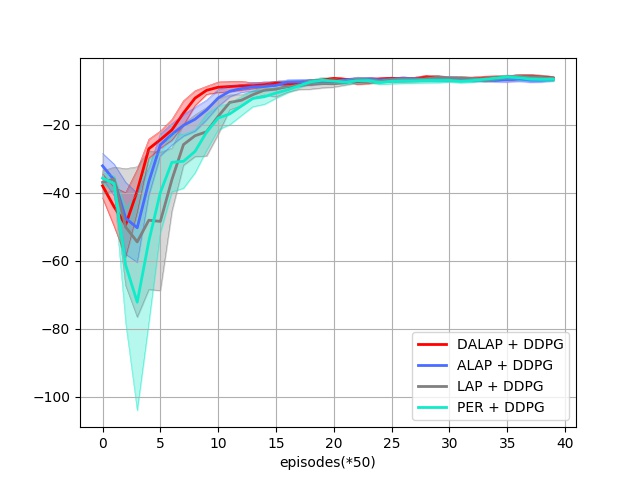}}
	\caption{Mean rewards of different training frameworks integrated with DDPG.}
	\label{DDPG}
\end{figure*}
	
Fig.\ref{DDPG} describes the average reward curves for different training frameworks integrated with DDPG, and DALAP exhibits the fastest convergence speed with the smallest training variance under all different mini-batch configurations. In the early stage of training, the confidence band of DALAP is much narrower than that of other algorithms,  which indicats DALAP is effective in suppressing the estimation bias due to PS.

\subsection{Results with MADDPG}\label{MADDPGresult}

The training framework carrier in \textbf{simple$\_$tag} is MADDPG. \textbf{simple$\_$tag} will follow the model network structure, hyperparameter configuration and experience pool volume setting in \textbf{simple}. Unlike DQN and DDPG, MADDPG is a multi-agent algorithm, and the agents in \textbf{simple$\_$tag} will be divided into two camps of a pursuit-escape relationship. Both pursuers and evaders adopt the zero-sum reward shaping, and the reward value is determined by the relative distance.

\begin{figure*}[!t]
	\centering
	\subfigure[batchsize=32]{
		\includegraphics[scale=0.355]{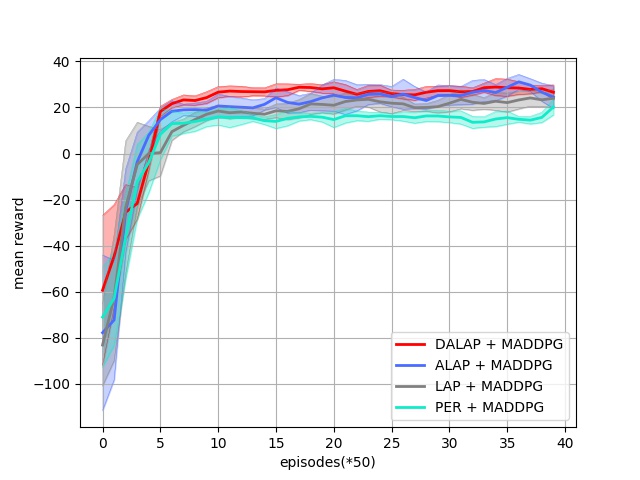}}
	\subfigure[batchsize=64]{
		\includegraphics[scale=0.355]{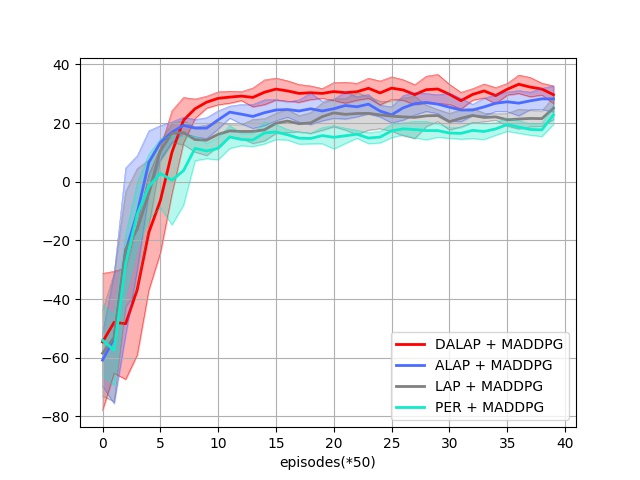}}
	\subfigure[batchsize=128]{
		\includegraphics[scale=0.355]{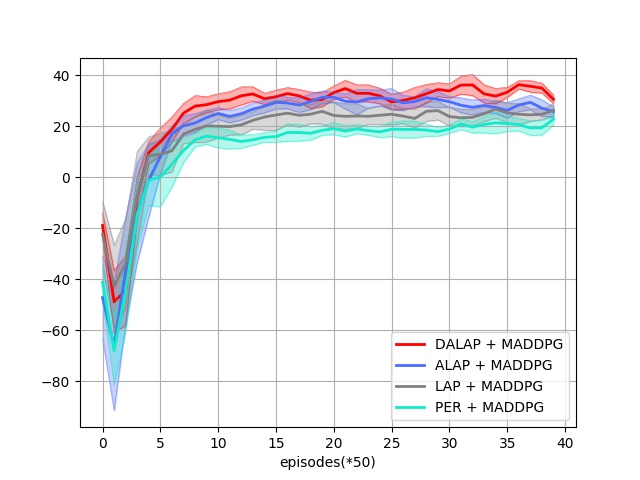}}
	\caption{Mean rewards of different training frameworks integrated with MADDPG.}
	\label{MADDPG}
\end{figure*}

Fig.\ref{MADDPG} describes the average reward curves for different training frameworks integrated with MADDPG. It is obvious that DALAP outperforms other algorithms under any mini-batch configurations. When mini-batch=64, although the variance of DALAP is equal to that of ALAP, the steady-state rewards of DALAP is much higher than the rest of the algorithms. In other cases, DALAP has the faster convergence speed, and the smaller training variance.

With the above experimental results, we have fully demonstrated the great advantages of DALAP in terms of convergence speed and stability. Meanwhile, DALAP can be applied to value-function based, policy-gradient based, and multi-agent algorithms, which shows its generality.

\section{CONCLUSIONS}\label{conclusion}

In this paper, we propose an advanced and generalized reinforcement learning training framework called DALAP, which greatly improves the convergence rate and stability of the algorithm. First of all, we theoreticaly prove the positive correlation between the estimation bias and the hyperparameter $\beta$. On this basis, we designed a PSAN network to directly quantify the change degree of the shifted state-action distribution, so as to estimate a more accurate $\beta$, and fundamentally address the estimation bias problem of PER. In addition, a Priority-Encouragement mechanism is proposed to screen out more high-quality transitions with learning value, which further improves the training speed. Finally, the superior performance of DALAP is verified by comparative tests.

\bibliographystyle{IEEEtran}

\vfill
\end{document}